\documentclass[11pt,letterpaper]{article}
\usepackage{naaclhlt2016}
\usepackage{times}
\usepackage{latexsym}
\usepackage{amssymb}
\usepackage{amsthm}
\usepackage{xcolor}
\usepackage{amsmath}
\usepackage{graphicx}
\usepackage{scalerel}
\usepackage{breqn}
\usepackage{url}

\DeclareMathOperator*{\Bigcdot}{\scalerel*{\cdot}{\bigodot}}

\newcommand\numberthis{\addtocounter{equation}{1}\tag{\theequation}}

\newcommand{\mfar}[1]{\textcolor{blue}{\bf\small [#1 --MFAR]}}

\newcommand{\ignore}[1]{}

\naaclfinalcopy 
\def\naaclpaperid{40} 

\title{Morphological Inflection Generation Using \\
Character Sequence to Sequence Learning}
\author{Manaal Faruqui$^1$ \quad Yulia Tsvetkov$^1$ \quad Graham Neubig$^2$ \quad Chris Dyer$^1$\\
$^1$Language Technologies Institute, Carnegie Mellon University, USA\\
$^2$Graduate School of Information Science, Nara Institute of Science and Technology, Japan\\
{\sf \{mfaruqui,ytsvetko,cdyer\}@cs.cmu.edu neubig@is.naist.jp}\\
}
\date{}

\begin{document}

\maketitle

\begin{abstract}
Morphological inflection generation is the task of generating the inflected
form of a given lemma corresponding to a particular linguistic
transformation. We model the problem of inflection generation as a character
sequence to sequence learning problem and present a variant of the neural
encoder-decoder model for solving it.
Our model is language independent and can be trained in both
supervised and semi-supervised settings. We evaluate our system on seven
datasets of morphologically rich languages and achieve either better or
comparable results to existing state-of-the-art models of inflection generation.
\end{abstract}

\section{Introduction}

Inflection is the word-formation mechanism to express different grammatical
categories such as tense, mood, voice, aspect, person, gender, number and case.
Inflectional morphology is often realized by the concatenation of bound morphemes (prefixes and suffixes) to a root form or stem, but nonconcatenative processes such as ablaut and infixation are found in many languages as well. 
Table~\ref{tab:inflex} shows the possible inflected forms of the German stem
\textit{Kalb} (calf) when it is used in different cases and numbers.  The inflected forms are the result of both ablaut (e.g., \textit{a}$\rightarrow$\textit{\"a}) and suffixation (e.g., +\textit{ern}).


Inflection generation is useful for reducing data sparsity in morphologically
complex languages.  For example, statistical machine translation suffers
from data sparsity when translating morphologically-rich languages,
since every surface form is considered an independent entity.
Translating into lemmas in the target language, and then
applying inflection generation as a post-processing step,
has been shown to alleviate the sparsity problem  \cite{minkov2007generating,toutanova2008applying,clifton:11,fraser:12,victor:13}.
Modeling inflection generation has also been used to improve language
modeling \cite{victor:13:lm}, identification of multi-word expressions
\cite{Oflazer:2004}, among other applications.

\begin{table}[!tb]
  \centering
  \begin{tabular}{lll}
  \hline
   & singular & plural \\
  \hline
  nominative & Kalb & K\"alber\\
  accusative & Kalb & K\"alber\\
  dative & Kalb & K\"albern\\
  genitive & Kalbes & K\"alber\\
  \hline
  \end{tabular}
  \caption{An example of an inflection table from the German noun dataset for
the word \textit{Kalb} (calf).}
  \label{tab:inflex}
\end{table}

The traditional approach to modeling inflection relies on hand-crafted finite
state transducers and lexicography, e.g., using \textit{two-level morphology} \cite{twolevel,kaplan1994regular}.
 Such systems are appealing since they correspond to linguistic theories, but they are expensive to create, they can be fragile \cite{oflazer1996error},
and the composed transducers can be impractically large. As an alternative,
machine learning models have been proposed to generate inflections
from root forms as string transduction
\cite{yarowsky:00,wicentowski:04,de:11,ddn:13,ahlberg:14,hulden:14,ahlberg:15,kondrak:15:inflection}. However, these impose either assumptions about the
set of possible morphological processes (e.g. affixation) or require careful
feature engineering.

In this paper, we present a model of inflection generation based on a neural
network sequence to sequence transducer. The root form is represented as
sequence of characters, and this is the input to an encoder-decoder architecture
\cite{cho:14:phrase,sutskever:14}. 
The model transforms its input to a sequence of
output characters representing the inflected form (\S\ref{sec:model}).
Our model makes no assumptions about morphological
processes, and our features are simply the individual characters. The
model is trained on pairs of root form and inflected forms obtained
from inflection tables extracted from Wiktionary.\footnote{\url{www.wiktionary.org}}
We improve the supervised model with unlabeled data, by integrating a
character language model trained on the vocabulary of the language.

Our experiments show that the model achieves better or comparable results to
state-of-the-art methods on the benchmark inflection generation tasks (\S\ref{sec:expts}).
For example, our model is able to learn long-range relations between character sequences in
the string aiding the inflection generation process required by Finnish vowel
harmony (\S\ref{sec:analysis}), which helps it obtain
the current best results in that language. We have publicly released our code for inflection generation.\footnote{\url{https://github.com/mfaruqui/morph-trans}}

\begin{figure}[tb]
  \centering
  \includegraphics[width=\columnwidth]{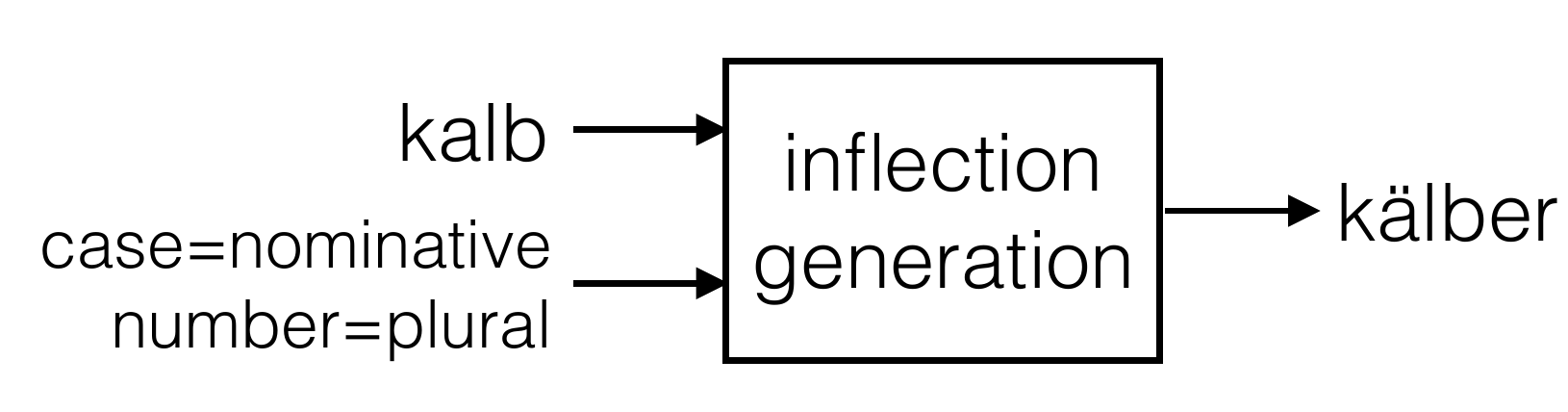}
  \caption{A general inflection generation model.}
  \label{fig:outline}
\end{figure}

\section{Inflection Generation: Background}
\label{sec:background}

\newcite{ddn:13} formulate the task of supervised inflection generation
for a given root form, based on a large number of training inflection tables
extracted from Wiktionary. Every
inflection table contains the inflected form of a given root word
corresponding to different linguistic transformations (cf.
Table~\ref{tab:inflex}).
Figure~\ref{fig:outline} shows the inflection generation framework.
Since the release of the Wiktionary dataset, several different models have
reported performance on this dataset. As we are also using this dataset, we
will now review these models.

We denote the models of \newcite{ddn:13}, \newcite{ahlberg:14},
\newcite{ahlberg:15}, and \newcite{kondrak:15:inflection}, by DDN13, AFH14,
AFH15, and NCK15 respectively. These models perform inflection generation
as string transduction and largely consist of three major components:
(1) Character alignment of word forms in a table;
(2) Extraction of string transformation rules;
(3) Application of rules to new root forms.

\begin{figure}[tb]
  \centering
  \includegraphics[width=0.75\columnwidth]{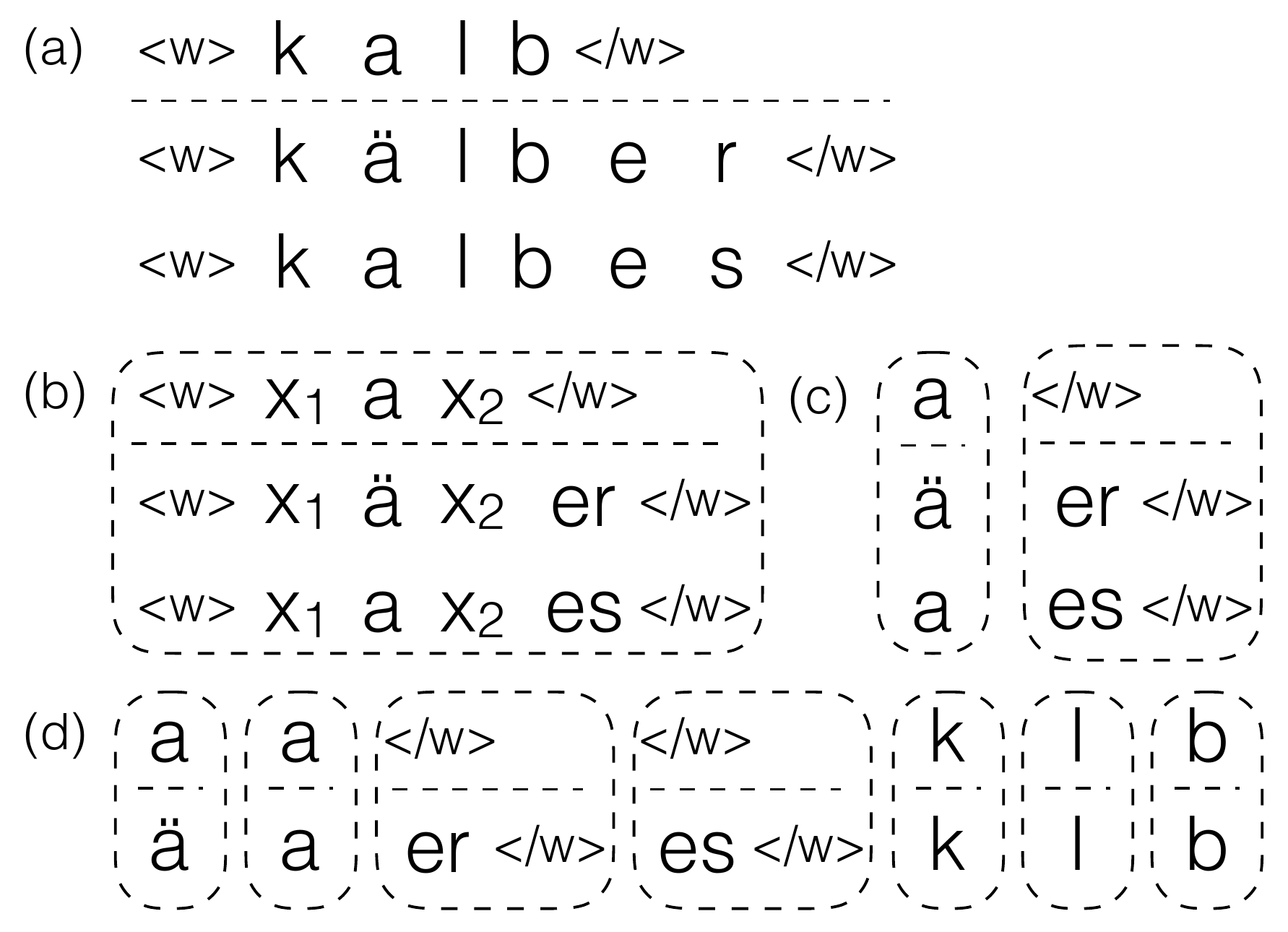}
  \caption{Rule extraction: (a) Character aligned-table;
(b) Table-level rule of AFH14, AFH15 (c) Vertical rules of DDN13 and
(d) Atomic rules of NCK15.}
  \label{fig:rules}
\end{figure}

The first step is learning character
alignments across inflected forms in a table. Figure~\ref{fig:rules} (a) shows
alignment between three word forms of \textit{Kalb}. Different models use
different heuristic algorithms for alignments such as edit distance,
dynamic edit distance \cite{eisner:02,oncina:06}, and longest subsequence
alignment \cite{lcs:00}. Aligning characters across word forms provide
spans of characters that have changed and spans that remain unchanged.
These spans are used to extract rules for inflection
generation for different inflection types as shown in Figure~\ref{fig:rules}
(b)--(d).

By applying the extracted rules to new root forms, inflected words can be
generated. DDN13 use a semi-Markov model \cite{sarawagi:04} to predict what
rules should be applied, using character $n$-grams ($n=1$ to $4$) as features.
AFH14 and AFH15 use substring features extracted from words to match an input
word to a rule table. NCK15 use a semi-Markov
model inspired by DDN13, but additionally use target $n$-grams and joint
$n$-grams as features sequences while selecting the rules.

\paragraph{Motivation for our model.} Morphology often makes references to
segmental features, like place or manner of articulation, or voicing status
\cite{chomsky68sound}.
While these can be encoded as features in existing work, our
approach treats segments as vectors of features ``natively''. Our approach
represents every character as a bundle of continuous features,
instead of using discrete surface character sequence features. Also, our model
uses features as part of the transduction rules themselves, whereas in existing
work features are only used to rescore rule applications.

In existing work, the
learner implicitly specifies the class of rules that can be learned, such as
``delete'' or ``concatenate''. To deal with phenomenona like segment
lengthening in English: \textit{run} $\rightarrow$ \textit{running}; or
reduplication in Hebrew: \textit{Kelev} $\rightarrow$ \textit{Klavlav},
\textit{Chatul} $\rightarrow$ \textit{Chataltul}; (or consonant gradation in
Finnish), where the affixes are induced from characters of the root form, one
must engineer a new rule class, which leads to poorer estimates due to data
sparsity. By modeling inflection generation as a task of generating
a character sequence, one character at a time, we do away with such problems.

\section{Neural Encoder-Decoder Models}
\label{sec:seqback}

Here, we describe briefly the underlying framework of our inflection generation
model, called the recurrent neural network (RNN) encoder-decoder
\cite{cho:14:phrase,sutskever:14}
which is used to transform an input sequence $\vec{x}$ to output sequence
$\vec{y}$.
We represent an item by $x$, a sequence of items by $\vec{x}$, vectors by
$\mathbf{x}$, matrices by $\mathbf{X}$, and sequences of vectors by
$\vec{\mathbf{x}}$.

\subsection{Formulation}
\label{sec:encdec}

In the encoder-decoder framework, an encoder reads a variable length input
sequence, 
a sequence of vectors $\vec{\mathbf{x}} = \langle \mathbf{x}_1,
\cdots, \mathbf{x}_T\rangle$ (corresponding to a sequence of input symbols
$\vec{x} = \langle x_1,
\cdots, x_T\rangle$) and generates a fixed-dimensional vector
representation of the sequence.
$\mathbf{x}_t \in \mathbb{R}^{l}$ is an input vector of length $l$.
The most common approach is to use an RNN such that:
\begin{equation}
\mathbf{h}_t = f(\mathbf{h}_{t-1}, \mathbf{x}_t)
\end{equation}
where $\mathbf{h}_t \in \mathbb{R}^n$ is a hidden state at time $t$, and
$f$ is generally a non-linear transformation, producing
$\mathbf{e} := \mathbf{h}_{T+1}$ as the input representation.
The decoder is
trained to predict the next output $y_t$ given the encoded input vector
$\mathbf{e}$ and all the previously predicted outputs
$\langle y_1, \cdots y_{t-1}\rangle$.
In other words, the decoder defines a probability over the output sequence
$\vec{y} = \langle y_1, \cdots, y_{T'}\rangle$ by decomposing
the joint probability into ordered conditionals:
\begin{equation}
p(\vec{y}|\vec{x}) = \prod_{t=1}^{T'} \nolimits p(y_t |
\mathbf{e}, \langle y_1, \cdots, y_{t-1}\rangle)
\end{equation}
With a decoder RNN, we can first obtain the hidden layer at time $t$ as:
$\mathbf{s}_t = g(\mathbf{s}_{t-1}, {\{\mathbf{e}, \mathbf{y}_{t-1}\}})$
and feed this into a softmax layer to obtain the conditional probability as:
\begin{equation}
p(y_t = i | \mathbf{e},\vec{y}_{<t}) =
\mathrm{softmax}(\mathbf{W}_s\mathbf{s}_t + \mathbf{b}_s)_i
\end{equation}
where, $\vec{y}_{<t} = \langle y_1, \cdots, y_{t-1}\rangle$.
In recent work, both $f$ and $g$ are generally LSTMs,
a kind of RNN which we describe next.

\subsection{Long Short-Term Memory (LSTM)}
\label{sec:lstm}

In principle, RNNs allow retaining information from time
steps in the distant past, but the nonlinear ``squashing'' functions applied in
the calculation of each $\mathbf{h}_t$ result in a decay of the error signal
used in training with backpropagation. LSTMs are a variant of RNNs designed to
cope with this ``vanishing gradient'' problem using an extra memory ``cell''
\cite{hochreiter:1997,graves:2013}.
Past work explains the computation within an LSTM through the metaphors of
deciding how much of the current input to pass into memory or forget.
We refer interested readers to the original papers for details.

\section{Inflection Generation Model}
\label{sec:model}

\begin{figure*}[tb]
  \centering
  \includegraphics[width=1.6\columnwidth]{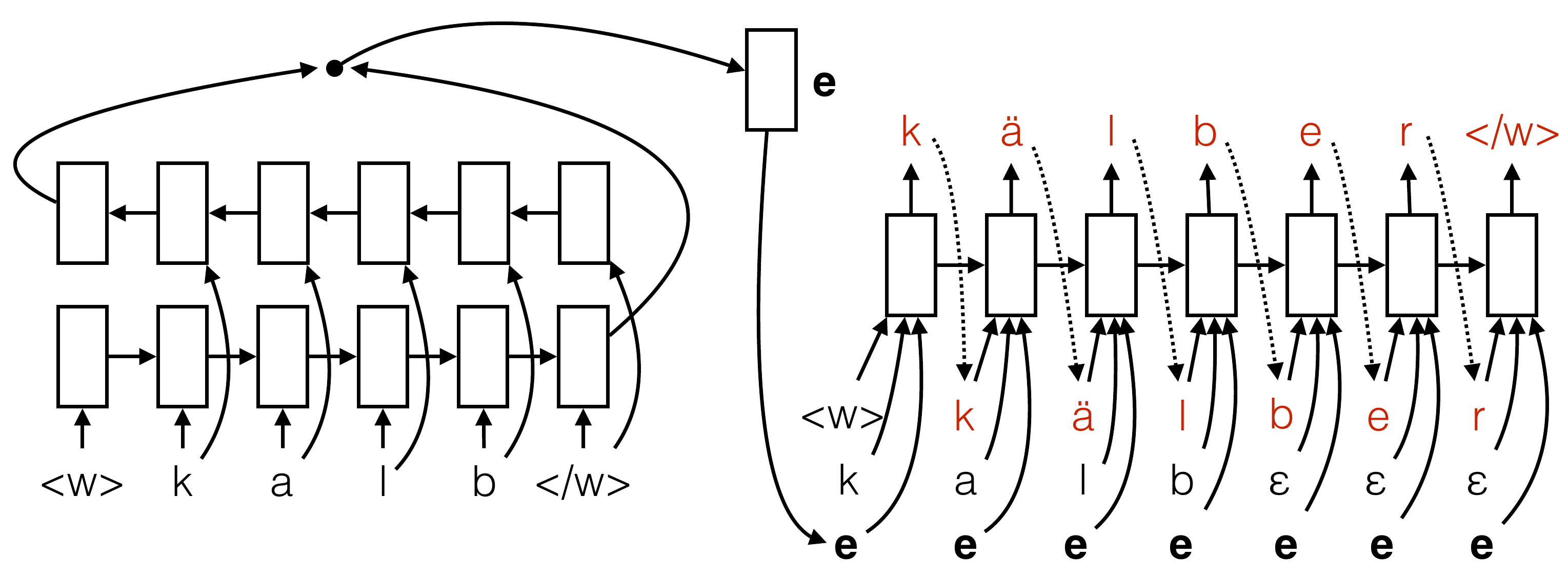}
  \caption{The modified encoder-decoder architecture for inflection generation. Input characters are shown in black and predicted
characters are shown in red. $\Bigcdot$ indicates the append operation.}
  \label{fig:model}
\end{figure*}

We frame the problem of inflection generation as a sequence to sequence
learning problem of character sequences.
The standard encoder-decoder models were designed for machine
translation where the objective is to translate a sentence (sequence of words)
from one language to a semantically equivalent sentence (sequence of words)
in another language.
We can easily port the encoder-decoder translation model for inflection
generation. Our model predicts the sequence
of characters in the inflected string given the characters in the root word
(input).

However, our problem differs from the above setting in two ways:
(1) the input and output character sequences are mostly similar except for
the inflections; (2) the input and output character sequences have different
semantics. Regarding the first difference, taking the word \textit{play} as an
example, the inflected forms corresponding to past tense and continuous forms
are \textit{played} and \textit{playing}.
To better use this correspondence between the input and output sequence, we
also feed the input sequence directly into the decoder:
\begin{equation}
\label{equ:hidden}
\mathbf{s}_t =
g(\mathbf{s}_{t-1}, \{\mathbf{e}, \mathbf{y}_{t-1}, \mathbf{x}_t\})
\end{equation}
where, $g$ is the decoder LSTM, and $\mathbf{x}_t$ and $\mathbf{y}_t$ are the
input and output character vectors respectively. Because the lengths of the
input and output sequences are not equal, we feed an $\epsilon$ character
in the decoder, indicating null input, once the input sequence runs out of
characters. These $\epsilon$ character vectors are parameters that are learned
by our model, exactly as other character vectors.

Regarding the second difference, to provide the model
the ability to learn the transformation of semantics from input to output, we
apply an affine transformation on the encoded vector $\mathbf{e}$:
\begin{equation}
\label{equ:affine}
\mathbf{e} \leftarrow \mathbf{W}_{trans}\mathbf{e} + \mathbf{b}_{trans}
\end{equation}
where, $\mathbf{W}_{trans}, \mathbf{b}_{trans}$ are the transformation
parameters. Also, in the encoder we use a bi-directional LSTM
\cite{graves2005bidirectional} instead of a
uni-directional LSTM, as it has been shown to capture the sequence information
more effectively \cite{wang:15,ballesteros:15,bahdanau2015}.
Our resultant inflection generation model is shown in
Figure~\ref{fig:model}.

\subsection{Supervised Learning}
\label{sec:supervised}
The parameters of our model are the set of character vectors,
the transformation parameters ($\mathbf{W}_{trans}, \mathbf{b}_{trans}$),
and the parameters of the encoder and decoder LSTMs (\S\ref{sec:lstm}).
We use negative log-likelihood of the output character sequence as the loss
function:
\begin{equation}
\label{equ:suploss}
- \mathrm{log} \,p(\vec{y}|\vec{x}) = - \sum_{t=1}^{T'} \nolimits
\mathrm{log} \,p(y_t | \mathbf{e}, \vec{y}_{<t})
\end{equation}
We minimize the loss using stochastic updates with
AdaDelta~\cite{zeiler2012adadelta}. This is our purely supervised model for
inflection generation and we evaluate it in two different settings as
established by previous work:

\paragraph{Factored Model.}
In the first setting, we learn a separate model for each
type of inflection independent of the other possible inflections. For example,
in case of German nouns, we learn 8, and for German verbs, we learn 27
individual encoder-decoder inflection models (cf. Table~\ref{tab:dataset}).
There is no parameter sharing across these models. We call these factored
models of inflection generation.

\paragraph{Joint Model.}
In the second setting, while learning a model for an inflection type, we also
use the information of how the lemma inflects across all other inflection
types i.e., the inflection table of a root form is used to learn different
inflection models. We model this, by having the same encoder in the
encoder-decoder model across all inflection models.\footnote{We also tried
having the same encoder and decoder across inflection types, with just the
transformation matrix being different (equ.~\ref{equ:affine}), and observed
consistently worse results.}
The encoder in our model is learning a representation of
the input character sequence. Because all inflection models take the same input
but produce different outputs, we hypothesize that having the same encoder
can lead to better estimates.

\subsection{Semi-supervised Learning}
\label{sec:semi}

The model we described so far relies entirely on the availability of pairs
of root form and inflected word form for learning to generate inflections.
Although such supervised models can be
used to obtain inflection generation models \cite{ddn:13,ahlberg:15}, it
has been shown that unlabeled data can generally improve the performance of
such systems \cite{ahlberg:14,kondrak:15:inflection}.
The vocabulary of the words of a language encode information about
what correct sequences of characters in a language
look like. Thus, we learn a language model over the character
sequences in a vocabulary extracted from a large unlabeled corpus. We
use this language model to make predictions about the next character in the
sequence given the previous characters, in following two settings.

\begin{table}[!tb]
  \centering
  \begin{tabular}{|c|c|}
  \hline
  $p_{\mathrm{LM}}(\vec{y})$ & $p(\vec{y}|\vec{x})$ \\
  len($\vec{y}$) - len($\vec{x}$) & levenshtein($\vec{y}$, $\vec{x}$) \\
  same-suffix($\vec{y}$, $\vec{x}$)? & subsequence($\vec{y}$, $\vec{x}$)? \\
  same-prefix($\vec{y}$, $\vec{x}$)? & subsequence($\vec{x}$, $\vec{y}$)? \\
  \hline
  \end{tabular}
  \caption{Features used to rerank the inflected outputs. $\vec{x}$,
$\vec{y}$ denote the root and inflected character sequences resp.}
  \label{tab:rerankfeat}
\end{table}

\paragraph{Output Reranking.} In the first setting, we first train the
inflection generation model using the supervised setting as described in
\S\ref{sec:supervised}. While making predictions for inflections, we use beam
search to generate possible output character sequences and rerank them using
the language model probability along with other easily extractable features as
described in Table~\ref{tab:rerankfeat}. We use pairwise ranking optimization
(PRO) to learn the reranking model \cite{pro}. The reranker is trained on the
beam output of dev set and evaluated on test set.

\paragraph{Language Model Interpolation.} In the second
setting, we interpolate the probability of observing the next character
according to the language model with the probability according to our inflection
generation model. Thus, the loss function becomes:
\begin{align*}
\label{equ:semiloss}
- \mathrm{log} \,p(\vec{y}|\vec{x}) = \frac{1}{Z}
\sum_{t=1}^{T'} \nolimits
&- \mathrm{log} \,p(y_t | \mathbf{e}, \vec{y}_{<t}) \\
  &- \lambda \mathrm{log} \, p_{\mathrm{LM}}(y_t |\vec{y}_{<t}) \numberthis
\end{align*}
where $p_{LM}(y_t |\vec{y}_{<t})$ is the
probability of observing the word $y_t$ given the history estimated according
to a language model, $\lambda \ge 0$ is the interpolation parameter
which is learned during training and $Z$ is the normalization factor. This
formulation lets us use any off-the-shelf pre-trained character language model
easily (details in \S\ref{sec:expts}).

\subsection{Ensembling}
Our loss functions (equ.~\ref{equ:suploss} \& \ref{equ:semiloss}) formulated
using a neural network architecture are non-convex in nature and are thus
difficult to optimize. It has been shown that taking an ensemble of models
which were initialized differently and trained independently leads to
improved performance \cite{hansen1990neural,collobert2011natural}.
Thus, for each model type used in this work, we report results obtained using
an ensemble of models. So, while decoding we compute the probability of
emitting a character as the product-of-experts of the individual models in
the ensemble:
$p_{ens}(y_t | \cdot) = \frac{1}{Z}\prod_{i=1}^{k} p_{i}(y_t | \cdot)^{\frac{1}{k}}$
where, 
$p_{i}(y_t | \cdot)$ is the probability according to $i$-th
model and $Z$ is the normalization factor.

\begin{table}[!tb]
  \centering
  \begin{tabular}{lrr}
  \hline
  Dataset & root forms & Infl. \\
  \hline
  German Nouns (DE-N) & 2764 & 8 \\
  German Verbs (DE-V) & 2027 & 27 \\
  Spanish Verbs (ES-V) & 4055 & 57 \\
  Finnish NN \& Adj. (FI-NA) & 6400 & 28 \\
  Finnish Verbs (FI-V) & 7249 & 53 \\
  Dutch Verbs (NL-V) & 11200 & 9 \\
  French Verbs (FR-V) & 6957 & 48 \\
  \hline
  \end{tabular}
  \caption{The number of root forms and types of inflections across datasets.}
  \label{tab:dataset}
\end{table}

\section{Experiments}
\label{sec:expts}
We now conduct experiments using the described models. Note that not all
previously published models present results on all settings, and thus we
compare our results to them wherever appropriate.



\paragraph{Hyperparameters.}
Across all models described in this paper, we use the following
hyperparameters. In both the encoder and decoder models
we use single layer LSTMs with the hidden vector of length $100$.
The length of character vectors is the size of character vocabulary
according to each dataset. The parameters are regularized with
$\ell_2$, with the regularization constant $10^{-5}$.\footnote{Using dropout
did not improve our results.} The number of models for ensembling are $k=5$.
Models are trained for at most $30$ epochs and the model with best result on
development set is selected.

\subsection{Data}
\newcite{ddn:13} published the Wiktionary inflection dataset with training,
development and test splits. The development
and test sets contain 200 inflection tables each and the training sets consist
of the remaining data. This dataset contains inflections for German, Finnish
and Spanish.
This dataset was further augmented by \cite{kondrak:15:inflection},
by adding Dutch verbs extracted from CELEX lexical database \cite{celex},
French verbs from Verbsite, an online French conjugation dictionary and Czech
nouns and verbs from the Prague Dependnecy Treebank \cite{HajicHladkaPajas2001}.
As the dataset for Czech contains many incomplete tables, we do not use it for
our experiments. These datasets come with pre-specified training/dev/test
splits, which we use(cf. Table~\ref{tab:dataset}).
For each of these sets, the training data is restricted to
80\% of the total inflection tables, with 10\% for development and 10\% for
testing.

For semi-supervised experiments, we train a 5-gram character language model
with Witten-Bell smoothing \cite{bell1990text} using the
SRILM toolkit \cite{srilm}.
We train the character language models
on the list of unique word types extracted from the Wikipedia dump for each
language after filtering out words with characters unseen in the inflection
generation training dataset. We obtained $\approx$2 million unique words for each
language.


\subsection{Results}

\begin{table}[!tb]
  \centering
  \begin{tabular}{lrrr}
  \hline
  & DDN13 & NCK15 & Ours \\
  \hline
  DE-V & 94.76 & \textbf{97.50} & 96.72 \\
  DE-N & 88.31 & \textbf{88.60} & 88.12 \\
  ES-V & 99.61 & 99.80 & \textbf{99.81} \\
  FI-V & 97.23 & \textbf{98.10} & 97.81 \\
  FI-NA & 92.14 & 93.00 & \textbf{95.44} \\
  NL-V & 90.50 & 96.10 & \textbf{96.71} \\
  FR-V & 98.80 & \textbf{99.20} & 98.82 \\
  \hline
  Avg. & 94.47 & 96.04 & \textbf{96.20} \\
  \hline
  \end{tabular}
  \caption{Individual form prediction accuracy for \textbf{factored supervised} models.}
  \label{tab:sup}
\end{table}

\paragraph{Supervised Models.}
The individual inflected form accuracy for the factored model
(\S\ref{sec:supervised}) is shown in  Table~\ref{tab:sup}.
Across datasets, we obtain either comparable or better results than NCK15 while
obtaining on average an accuracy of $96.20$\% which is higher than both DDN13
and NCK15. Our factored model performs better than DDN13 and
NCK15 on datasets with large training set (ES-V, FI-V, FI-NA, NL-V, FR-V)
as opposed to datasets with small training set (DE-N, DE-V).
In the joint model setting (cf. Table~\ref{tab:joint}), on average,
we perform better than DDN13 and
AFH14 but are behind AFH15 by $0.11$\%. Our model improves in performance over
our factored model for DE-N, DE-V, and ES-V, which
are the three smallest training datasets. Thus, parameter sharing across
different inflection types helps the low-resourced scenarios.\footnote{Although NCK15 provide results in the joint model setting, they also use raw data in
the joint model which makes it incomparable to our model and other previous
models.}

\begin{table}[!tb]
  \centering
  \begin{tabular}{lrrrr}
  \hline
   & DDN13 & AFH14 & AFH15 & Ours \\
  \hline
  DE-V & 96.19 & 97.01 & \textbf{98.11} & 97.25 \\
  DE-N & 88.94 & 87.81 & \textbf{89.88} & 88.37\\
  ES-V & 99.67 & 99.52 & \textbf{99.92} & 99.86 \\
  FI-V & 96.43 & 96.36 & 97.14 & \textbf{97.97}\\
  FI-NA & 93.41 & 91.91 & 93.68 & \textbf{94.71}\\
  \hline
  Avg. & 94.93 & 94.53 & \textbf{95.74} & 95.63 \\
  \hline
  NL-V & 93.88 & -- & -- & 96.16\\
  FR-V & 98.60 & -- & -- & 98.74\\
  \hline
  Avg. & 95.30 & -- & -- & \textbf{96.15}\\
  \hline
  \end{tabular}
  \caption{Individual form prediction accuracy for \textbf{joint supervised} models.}
  \label{tab:joint}
\end{table}

\paragraph{Semi-supervised Models.}
We now evaluate the utility of character language models in inflection
generation, in two different settings as described earlier (\S\ref{sec:semi}).
We use the factored model as our base model in the following experiments as it
performed better than the joint model (cf. Table \ref{tab:sup} \&
\ref{tab:joint}). Our reranking model which uses the character language model
along with other features (cf. Table~\ref{tab:rerankfeat}) to select the best
answer from a beam of predictions, improves over almost all the datasets with
respect to the supervised model and is equal on average to AFH14 and NCK15
semi-supervised models with $96.45\%$ accuracy. We obtain the best reported
results on ES-V and FI-NA datasets (99.94\% and 95.66\% respectively).
However, our second semi-supervised model, the interpolation model, on average
obtains $96.08\%$ and is surprisingly worse than our supervised model
($96.20\%$).

\begin{table}[!tb]
  \centering
  \begin{tabular}{lrrrr}
  \hline
   & AFH14 & NCK15 & Interpol & Rerank \\
  \hline
  DE-V & 97.87 & \textbf{97.90} & 96.79 & 97.11 \\
  DE-N & 91.81 & \textbf{89.90} & 88.31 & 89.31 \\
  ES-V & 99.58 & 99.90 & 99.78 & \textbf{99.94} \\
  FI-V & 96.63 & \textbf{98.10} & 96.66 & 97.62 \\
  FI-NA & 93.82 & 93.60 & 94.60 & \textbf{95.66} \\
  \hline
  Avg. & \textbf{95.93} & 95.88 & 95.42 & \textbf{95.93} \\
  \hline
  NL-V & -- & 96.60 & \textbf{96.66} & 96.64 \\
  FR-V & -- & \textbf{99.20} & 98.81 & 98.94 \\
  \hline
  Avg. & -- & \textbf{96.45} & 96.08 & \textbf{96.45} \\
  \hline
  \end{tabular}
  \caption{Individual form prediction accuracy for \textbf{factored semi-supervised} models.}
  \label{tab:semi}
\end{table}

\begin{table}[!tb]
  \centering
  \begin{tabular}{lr}
  \hline
  Model & Accuracy \\
  \hline
  Encoder-Decoder & 79.08 \\
  Encoder-Decoder Attention & 95.64 \\
  Ours W/O Encoder & 84.04 \\
  \hline
  Ours & \textbf{96.20}\\
  \hline
  \end{tabular}
  \caption{Avg. accuracy across datasets of the encoder-decoder, attentional
encoder-decoder \& our model without encoder.}
  \label{tab:comp}
\end{table}

\paragraph{Comparison to Other Architectures.} Finally it is of interest how
our proposed model compares to more traditional neural models.
We compare our model against a standard encoder-decoder model, and an
encoder-decoder model with attention, both trained on root form
to inflected form character sequences. In a standard encoder-decoder
model \cite{sutskever:14}, the encoded input sequence vector is fed into the
hidden layer of the decoder as input, and is not available at every time step
in contrast to our model, where we additionally feed in $\mathbf{x}_t$ at every
time step as in equ.~\ref{equ:hidden}. An attentional model
computes a weighted average of the hidden layer of the input sequence, which is
then used along with the decoder hidden layer to make a prediction
\cite{bahdanau2015}. These models also do not take the root form character
sequence as inputs to the decoder. We also evaluate the utility of having an
encoder which computes a representation of the input character sequence in a
vector $\mathbf{e}$ by removing the encoder from our model in
Figure~\ref{fig:model}. The results in Table~\ref{tab:comp} show that we
outperform the encoder-decoder model, and the model without an encoder
substantially. Our model is slightly better than the attentional encoder-decoder
model, and is simpler as it does not have the additional attention layer.

\section{Analysis}
\label{sec:analysis}

\begin{figure}[tb]
  \centering
  \includegraphics[width=0.9\columnwidth]{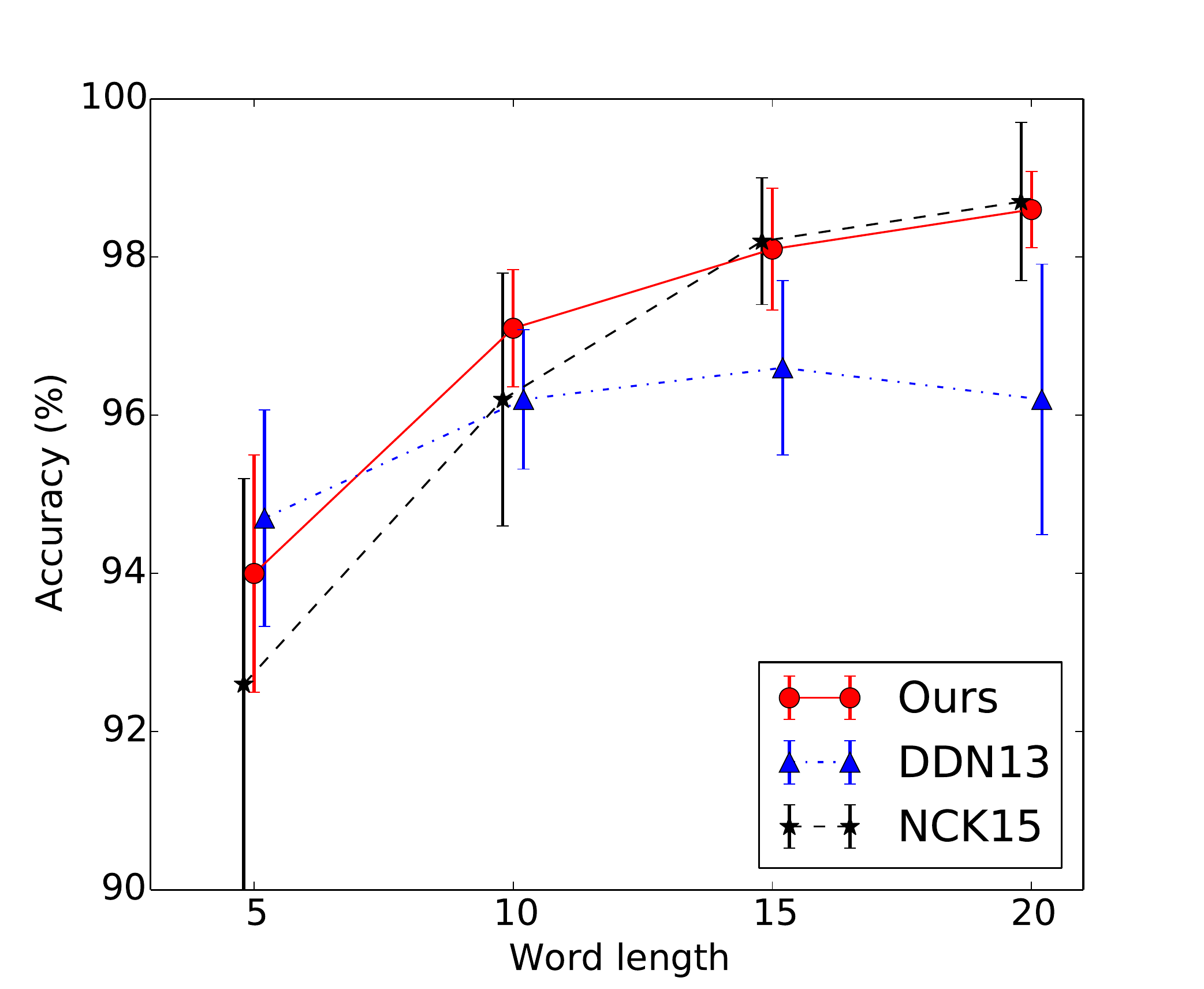}
  \caption{Plot of inflection prediction accuracy against the length of gold
inflected forms. The points are shown with minor offset along the x-axis to
enhance clarity.}
  \label{fig:length}
\end{figure}

\paragraph{Length of Inflected Forms.}
In Figure~\ref{fig:length} we show how the prediction accuracy of an inflected
form varies with respect to the length of the correct inflected
form.
To get stable
estimates, we bin the inflected forms according to their length:
$<5$, $[5, 10)$, $[10, 15)$, and $\ge 15$. The accuracy for each
bin is macro-averaged across 6 datasets\footnote{We remove DE-N as its
the smallest and shows high variance in results.} for our factored model and
the best models of DDN13 and NCK15.
Our model consistently shows improvement in performance as word length
increases and is significantly better than DDN13 on words of length more than
20 and is approximately equal to NCK15.
On words of length $<5$, we perform worse than DDN13
but better than NCK15. On average, our model has the least error margin across
bins of different word length as compared to both DDN13 and NCK15.
Using LSTMs in our model helps us make better predictions
for long sequences, since they have the ability to capture long-range
dependencies.

\begin{figure}[tb]
  \centering
  \includegraphics[width=0.9\columnwidth]{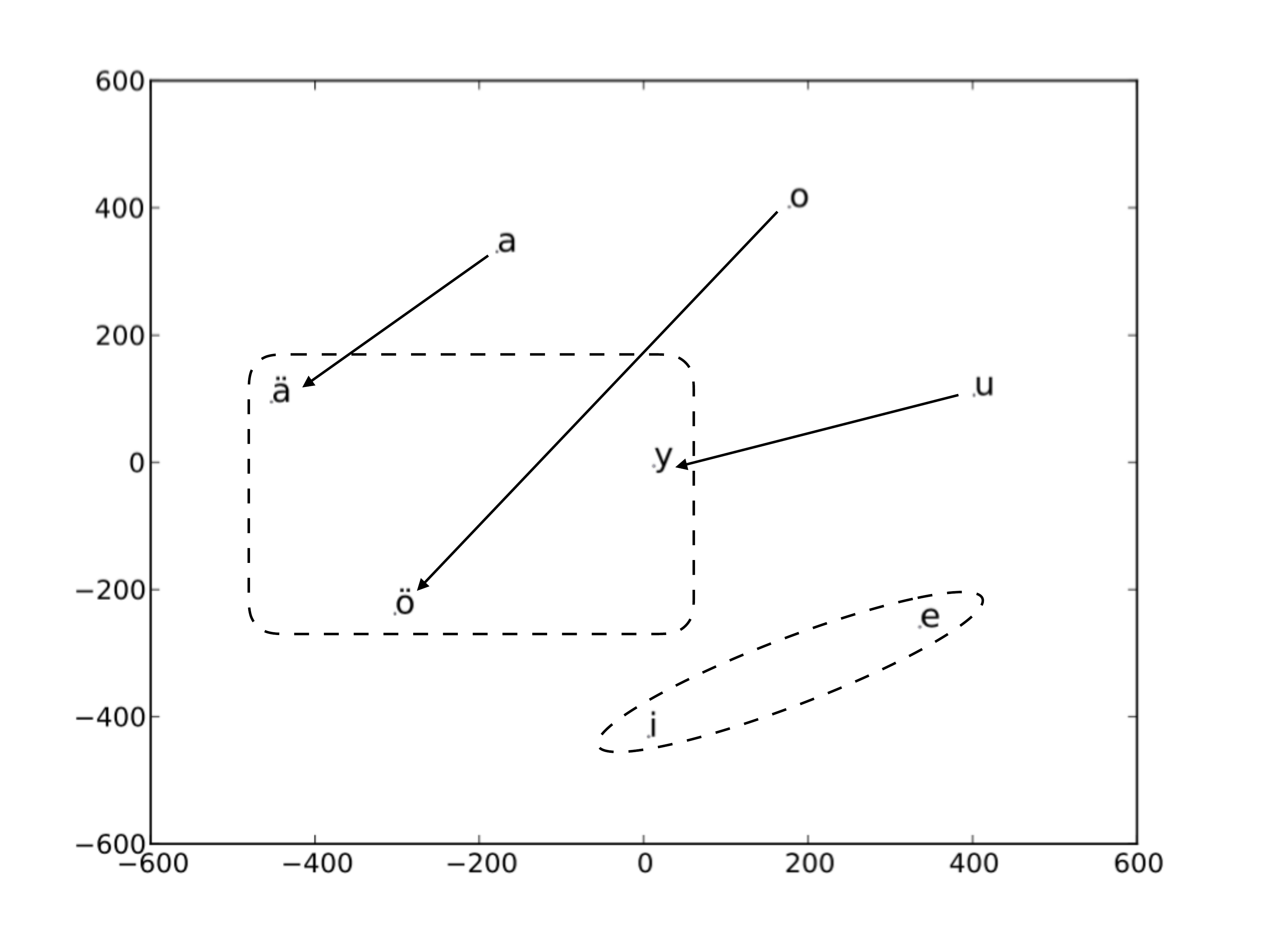}
  \caption{Plot of character vectors of Finnish vowels. Their organization shows that front, back and neutral vowel groups have been discovered. The
arrows show back and front vowel correspondences.}
  \label{fig:harmony}
\end{figure}

\paragraph{Finnish Vowel Harmony.}
Our model obtains the current best result on the Finnish noun and adjective
dataset, this dataset has the longest inflected words, some of which are $>30$
characters long. Finnish exhibits \textit{vowel harmony}, i.e, the occurrence
of a vowel is controlled by other vowels in the word. Finnish vowels are
divided into three groups: front (\"a, \"o, y), back (a, o, u), and neutral
(e, i). If back vowels are present in a stem, then the harmony is back (i.e,
front vowels will be absent), else the harmony is front (i.e, back vowels will
be absent). In compound words the suffix harmony is determined by
the final stem in the compound.
For example, our model correctly inflects the word \textit{fasisti} (fascist)
to obtain \textit{fasisteissa} and the compound
\textit{t\"arkkelyspitoinen} (starch containing) to
\textit{t\"arkkelyspitoisissa}.
The ability of our model to learn such relations between these
vowels helps capture vowel harmony. For FI-NA, our model obtains $99.87\%$ for
correctly predicting vowel harmony, and NCK15 obtains
$98.50\%$.
We plot the character vectors of these Finnish vowels (cf.
Figure~\ref{fig:harmony}) using t-SNE projection \cite{tsne} and observe
that the vowels are correctly grouped with visible transition from the back to
the front vowels.

\ignore{
\paragraph{Application in Language Generation.} Our inflection generation system
can be used in a natural language generation (NLG) system. \newcite{generation}
present an NLG system, which produces sentences with lemmas instead of
inflected word forms. We conduct an oracle experiment to check how much we can
improve this generation system if we are given the correct lemma and
morphological tags. We use our inflection generation system trained for spanish
verbs, to generate inflected verbs in their test set obtained from the
AnCora-UPF treebank \cite{ancoraupf}. The test set contains around $9100$ words,
out of which only $653$ are verbs, and which we inflect. The BLEU score
\cite{bleu} between sentences containing only lemmas and sentences containing
fully inflected words is $47.24$. When we inflect verbs the BLEU score improves
to $47.76$.
}

\section{Related Work}

Similar to the encoder in our framework,
\newcite{rastogi:2016} extract sub-word features using a
forward-backward LSTM from a word, and use them in a traditional weighted FST
to generate inflected forms. Neural encoder-decoder models of string
transduction have also been used for sub-word level transformations like
grapheme-to-phoneme conversion \cite{g2p,rao2015grapheme}.

Generation of inflectional morphology has been particularly useful in
statistical machine translation, both in translation from
morphologically rich languages \cite{goldwater2005improving}, and
into  morphologically rich languages
\cite{minkov2007generating,toutanova2008applying,clifton:11,fraser:12}. Modeling the morphological structure of a word has also shown to improve the
quality of word clusters \cite{clark:2003} and word vector representations
\cite{cotterell:2015}.

Inflection generation is complementary to the task of morphological
and phonological segmentation, where the existing word form needs to be
segmented to obtained meaningful sub-word units
\cite{creutz2005unsupervised,Snyder08unsupervisedmultilingual,poon2009unsupervised,narasimhan2015unsupervised,cotterell:phon,cotterell:2016}.
An additional line of work that benefits from implicit modeling
of morphology is neural character-based natural language processing,
e.g., part-of-speech tagging \cite{santos:14,wang:15} and dependency parsing
\cite{ballesteros:15}. These models have been successful when
applied to morphologically rich languages, as they are able to capture word
formation patterns.

\section{Conclusion}

We have presented a model that generates inflected forms of a given root form
using a neural network sequence to sequence string transducer. Our model obtains
state-of-the-art results and performs at par or better than existing inflection
generation models on seven different datasets.
Our model is able to learn long-range dependencies within
character sequences for inflection generation which makes it specially
suitable for morphologically rich languages.

\section*{Acknowledgements}
We thank Mans Hulden for help in explaining Finnish vowel harmony, and Garrett
Nicolai for making the output of his system available for comparison.
This work was sponsored in part by the National Science Foundation through award IIS-1526745.

\bibliography{naaclhlt2016}

\begin{thebibliography}{}

\bibitem[\protect\citename{Ahlberg \bgroup et al.\egroup }2014]{ahlberg:14}
Malin Ahlberg, Markus Forsberg, and Mans Hulden.
\newblock 2014.
\newblock Semi-supervised learning of morphological paradigms and lexicons.
\newblock In {\em Proc. of EACL}.

\bibitem[\protect\citename{Ahlberg \bgroup et al.\egroup }2015]{ahlberg:15}
Malin Ahlberg, Markus Forsberg, and Mans Hulden.
\newblock 2015.
\newblock Paradigm classification in supervised learning of morphology.
\newblock {\em Proc. of NAACL}.

\bibitem[\protect\citename{Baayen \bgroup et al.\egroup }1995]{celex}
Harald~R. Baayen, Richard Piepenbrock, and Leon Gulikers.
\newblock 1995.
\newblock {\em The {CELEX} Lexical Database. Release 2 ({CD-ROM})}.
\newblock {LDC}, University of Pennsylvania.

\bibitem[\protect\citename{Bahdanau \bgroup et al.\egroup }2015]{bahdanau2015}
Dzmitry Bahdanau, Kyunghyun Cho, and Yoshua Bengio.
\newblock 2015.
\newblock Neural machine translation by jointly learning to align and
  translate.
\newblock In {\em Proc. of ICLR}.

\bibitem[\protect\citename{Ballesteros \bgroup et al.\egroup
  }2015]{ballesteros:15}
Miguel Ballesteros, Chris Dyer, and Noah~A. Smith.
\newblock 2015.
\newblock Improved transition-based parsing by modeling characters instead of
  words with lstms.
\newblock In {\em Proc. of EMNLP}.

\bibitem[\protect\citename{Bell \bgroup et al.\egroup }1990]{bell1990text}
Timothy~C Bell, John~G Cleary, and Ian~H Witten.
\newblock 1990.
\newblock {\em Text compression}.
\newblock Prentice-Hall, Inc.

\bibitem[\protect\citename{Bergroth \bgroup et al.\egroup }2000]{lcs:00}
Lasse Bergroth, Harri Hakonen, and Timo Raita.
\newblock 2000.
\newblock A survey of longest common subsequence algorithms.
\newblock In {\em Proc. of SPIRE}.

\bibitem[\protect\citename{Chahuneau \bgroup et al.\egroup }2013a]{victor:13}
Victor Chahuneau, Eva Schlinger, Noah~A. Smith, and Chris Dyer.
\newblock 2013a.
\newblock Translating into morphologically rich languages with synthetic
  phrases.
\newblock In {\em Proc. of EMNLP}.

\bibitem[\protect\citename{Chahuneau \bgroup et al.\egroup
  }2013b]{victor:13:lm}
Victor Chahuneau, Noah~A Smith, and Chris Dyer.
\newblock 2013b.
\newblock Knowledge-rich morphological priors for bayesian language models.
\newblock In {\em Proc. of NAACL}.

\bibitem[\protect\citename{Cho \bgroup et al.\egroup }2014]{cho:14:phrase}
Kyunghyun Cho, Bart van Merrienboer, Caglar Gulcehre, Dzmitry Bahdanau, Fethi
  Bougares, Holger Schwenk, and Yoshua Bengio.
\newblock 2014.
\newblock Learning phrase representations using rnn encoder--decoder for
  statistical machine translation.
\newblock In {\em Proc. of EMNLP}.

\bibitem[\protect\citename{Chomsky and Halle}1968]{chomsky68sound}
N.~Chomsky and M.~Halle.
\newblock 1968.
\newblock {\em The Sound Pattern of English}.
\newblock Harper \& Row, New York, NY.

\bibitem[\protect\citename{Clark}2003]{clark:2003}
Alexander Clark.
\newblock 2003.
\newblock Combining distributional and morphological information for part of
  speech induction.
\newblock In {\em Proc. of EACL}.

\bibitem[\protect\citename{Clifton and Sarkar}2011]{clifton:11}
Ann Clifton and Anoop Sarkar.
\newblock 2011.
\newblock Combining morpheme-based machine translation with post-processing
  morpheme prediction.
\newblock In {\em Proc. of ACL}.

\bibitem[\protect\citename{Collobert \bgroup et al.\egroup
  }2011]{collobert2011natural}
Ronan Collobert, Jason Weston, L{\'e}on Bottou, Michael Karlen, Koray
  Kavukcuoglu, and Pavel Kuksa.
\newblock 2011.
\newblock Natural language processing (almost) from scratch.
\newblock {\em The Journal of Machine Learning Research}, 12:2493--2537.

\bibitem[\protect\citename{Cotterell and Sch\"{u}tze}2015]{cotterell:2015}
Ryan Cotterell and Hinrich Sch\"{u}tze.
\newblock 2015.
\newblock Morphological word-embeddings.
\newblock In {\em Proc. of NAACL}.

\bibitem[\protect\citename{Cotterell \bgroup et al.\egroup
  }2015]{cotterell:phon}
Ryan Cotterell, Nanyun Peng, and Jason Eisner.
\newblock 2015.
\newblock Modeling word forms using latent underlying morphs and phonology.
\newblock {\em Transactions of the Association for Computational Linguistics},
  3:433--447.

\bibitem[\protect\citename{Cotterell \bgroup et al.\egroup
  }2016]{cotterell:2016}
Ryan Cotterell, Tim Vieria, and Hinrich Sch\"utze.
\newblock 2016.
\newblock A joint model of orthography and morphological segmentation.
\newblock In {\em Proc. of NAACL}.

\bibitem[\protect\citename{Creutz and Lagus}2005]{creutz2005unsupervised}
Mathias Creutz and Krista Lagus.
\newblock 2005.
\newblock {\em Unsupervised morpheme segmentation and morphology induction from
  text corpora using Morfessor 1.0}.
\newblock Helsinki University of Technology.

\bibitem[\protect\citename{Dreyer and Eisner}2011]{de:11}
Markus Dreyer and Jason Eisner.
\newblock 2011.
\newblock Discovering morphological paradigms from plain text using a dirichlet
  process mixture model.
\newblock In {\em Proc. of EMNLP}.

\bibitem[\protect\citename{Durrett and DeNero}2013]{ddn:13}
Greg Durrett and John DeNero.
\newblock 2013.
\newblock Supervised learning of complete morphological paradigms.
\newblock In {\em Proc. of NAACL}.

\bibitem[\protect\citename{Eisner}2002]{eisner:02}
Jason Eisner.
\newblock 2002.
\newblock Parameter estimation for probabilistic finite-state transducers.
\newblock In {\em Proc. of ACL}.

\bibitem[\protect\citename{Fraser \bgroup et al.\egroup }2012]{fraser:12}
Alexander Fraser, Marion Weller, Aoife Cahill, and Fabienne Cap.
\newblock 2012.
\newblock Modeling inflection and word-formation in {SMT}.
\newblock In {\em Proc. of EACL}.

\bibitem[\protect\citename{Goldwater and McClosky}2005]{goldwater2005improving}
Sharon Goldwater and David McClosky.
\newblock 2005.
\newblock Improving statistical {MT} through morphological analysis.
\newblock In {\em Proc. of EMNLP}, pages 676--683.

\bibitem[\protect\citename{Graves \bgroup et al.\egroup
  }2005]{graves2005bidirectional}
Alex Graves, Santiago Fern{\'a}ndez, and J{\"u}rgen Schmidhuber.
\newblock 2005.
\newblock Bidirectional lstm networks for improved phoneme classification and
  recognition.
\newblock In {\em Proc. of ICANN}.

\bibitem[\protect\citename{Graves}2013]{graves:2013}
Alex Graves.
\newblock 2013.
\newblock Generating sequences with recurrent neural networks.
\newblock {\em CoRR}, abs/1308.0850.

\bibitem[\protect\citename{Haji\v{c} \bgroup et al.\egroup
  }2001]{HajicHladkaPajas2001}
Jan Haji\v{c}, Barbora Vidov\'{a}-Hladk\'{a}, and Petr Pajas.
\newblock 2001.
\newblock The {Prague Dependency Treebank}: Annotation structure and support.
\newblock In {\em Proc. of the IRCS Workshop on Linguistic Databases}.

\bibitem[\protect\citename{Hansen and Salamon}1990]{hansen1990neural}
Lars~Kai Hansen and Peter Salamon.
\newblock 1990.
\newblock Neural network ensembles.
\newblock In {\em Proc. of PAMI}.

\bibitem[\protect\citename{Hochreiter and Schmidhuber}1997]{hochreiter:1997}
Sepp Hochreiter and J\"urgen Schmidhuber.
\newblock 1997.
\newblock Long short-term memory.
\newblock {\em Neural Computation}, 9(8):1735--1780.

\bibitem[\protect\citename{Hopkins and May}2011]{pro}
Mark Hopkins and Jonathan May.
\newblock 2011.
\newblock Tuning as ranking.
\newblock In {\em Proc. of EMNLP}.

\bibitem[\protect\citename{Hulden}2014]{hulden:14}
Mans Hulden.
\newblock 2014.
\newblock Generalizing inflection tables into paradigms with finite state
  operations.
\newblock In {\em Proc. of the Joint Meeting of {SIGMORPHON} and {SIGFSM}}.

\bibitem[\protect\citename{Kaplan and Kay}1994]{kaplan1994regular}
Ronald~M Kaplan and Martin Kay.
\newblock 1994.
\newblock Regular models of phonological rule systems.
\newblock {\em Computational linguistics}, 20(3):331--378.

\bibitem[\protect\citename{Koskenniemi}1983]{twolevel}
Kimmo Koskenniemi.
\newblock 1983.
\newblock Two-level morphology: A general computational model for word-form
  recognition and production.
\newblock {\em University of Helsinki}.

\bibitem[\protect\citename{Ling \bgroup et al.\egroup }2015]{wang:15}
Wang Ling, Tiago Lu{\'\i}s, Lu{\'\i}s Marujo, R\'{a}mon~Fernandez Astudillo,
  Silvio Amir, Chris Dyer, Alan~W Black, and Isabel Trancoso.
\newblock 2015.
\newblock Finding function in form: Compositional character models for open
  vocabulary word representation.
\newblock In {\em Proc. of EMNLP}.

\bibitem[\protect\citename{Minkov \bgroup et al.\egroup
  }2007]{minkov2007generating}
Einat Minkov, Kristina Toutanova, and Hisami Suzuki.
\newblock 2007.
\newblock Generating complex morphology for machine translation.
\newblock In {\em Proc. of ACL}.

\bibitem[\protect\citename{Narasimhan \bgroup et al.\egroup
  }2015]{narasimhan2015unsupervised}
Karthik Narasimhan, Regina Barzilay, and Tommi Jaakkola.
\newblock 2015.
\newblock An unsupervised method for uncovering morphological chains.
\newblock {\em TACL}.

\bibitem[\protect\citename{Nicolai \bgroup et al.\egroup
  }2015]{kondrak:15:inflection}
Garrett Nicolai, Colin Cherry, and Grzegorz Kondrak.
\newblock 2015.
\newblock Inflection generation as discriminative string transduction.
\newblock In {\em Proc. of NAACL}.

\bibitem[\protect\citename{Oflazer \bgroup et al.\egroup }2004]{Oflazer:2004}
Kemal Oflazer, \"{O}zlem \c{c}etino\u{g}lu, and Bilge Say.
\newblock 2004.
\newblock Integrating morphology with multi-word expression processing in
  turkish.
\newblock In {\em Proc. of the Workshop on Multiword Expressions}.

\bibitem[\protect\citename{Oflazer}1996]{oflazer1996error}
Kemal Oflazer.
\newblock 1996.
\newblock Error-tolerant finite-state recognition with applications to
  morphological analysis and spelling correction.
\newblock {\em Computational Linguistics}, 22(1):73--89.

\bibitem[\protect\citename{Oncina and Sebban}2006]{oncina:06}
Jose Oncina and Marc Sebban.
\newblock 2006.
\newblock Learning stochastic edit distance: Application in handwritten
  character recognition.
\newblock {\em Pattern recognition}, 39(9):1575--1587.

\bibitem[\protect\citename{Poon \bgroup et al.\egroup
  }2009]{poon2009unsupervised}
Hoifung Poon, Colin Cherry, and Kristina Toutanova.
\newblock 2009.
\newblock Unsupervised morphological segmentation with log-linear models.
\newblock In {\em Proc. of NAACL}.

\bibitem[\protect\citename{Rao \bgroup et al.\egroup }2015]{rao2015grapheme}
Kanishka Rao, Fuchun Peng, Hasim Sak, and Fran{\c{c}}oise Beaufays.
\newblock 2015.
\newblock Grapheme-to-phoneme conversion using long short-term memory recurrent
  neural networks.
\newblock In {\em Proc. of ICASSP}.

\bibitem[\protect\citename{Rastogi \bgroup et al.\egroup }2016]{rastogi:2016}
Pushpendre Rastogi, Ryan Cotterell, and Jason Eisner.
\newblock 2016.
\newblock Weighting finite-state transductions with neural context.
\newblock In {\em Proc. of NAACL}.

\bibitem[\protect\citename{Santos and Zadrozny}2014]{santos:14}
Cicero~D. Santos and Bianca Zadrozny.
\newblock 2014.
\newblock Learning character-level representations for part-of-speech tagging.
\newblock In {\em Proc. of ICML}.

\bibitem[\protect\citename{Sarawagi and Cohen}2004]{sarawagi:04}
Sunita Sarawagi and William~W Cohen.
\newblock 2004.
\newblock Semi-markov conditional random fields for information extraction.
\newblock In {\em Proc. of NIPS}.

\bibitem[\protect\citename{Snyder and
  Barzilay}2008]{Snyder08unsupervisedmultilingual}
Benjamin Snyder and Regina Barzilay.
\newblock 2008.
\newblock Unsupervised multilingual learning for morphological segmentation.
\newblock In {\em In The Annual Conference of the}.

\bibitem[\protect\citename{Stolcke}2002]{srilm}
Andreas Stolcke.
\newblock 2002.
\newblock Srilm-an extensible language modeling toolkit.
\newblock In {\em Proc. of Interspeech}.

\bibitem[\protect\citename{Sutskever \bgroup et al.\egroup }2014]{sutskever:14}
Ilya Sutskever, Oriol Vinyals, and Quoc~VV Le.
\newblock 2014.
\newblock Sequence to sequence learning with neural networks.
\newblock In {\em Proc. of NIPS}.

\bibitem[\protect\citename{Toutanova \bgroup et al.\egroup
  }2008]{toutanova2008applying}
Kristina Toutanova, Hisami Suzuki, and Achim Ruopp.
\newblock 2008.
\newblock Applying morphology generation models to machine translation.
\newblock In {\em Proc. of ACL}, pages 514--522.

\bibitem[\protect\citename{van~der Maaten and Hinton}2008]{tsne}
Laurens van~der Maaten and Geoffrey Hinton.
\newblock 2008.
\newblock {Visualizing Data using t-SNE}.
\newblock {\em Journal of Machine Learning Research}, 9:2579--2605.

\bibitem[\protect\citename{Wicentowski}2004]{wicentowski:04}
Richard Wicentowski.
\newblock 2004.
\newblock Multilingual noise-robust supervised morphological analysis using the
  wordframe model.
\newblock In {\em Proc. of SIGPHON}.

\bibitem[\protect\citename{Yao and Zweig}2015]{g2p}
Kaisheng Yao and Geoffrey Zweig.
\newblock 2015.
\newblock Sequence-to-sequence neural net models for grapheme-to-phoneme
  conversion.
\newblock In {\em Proc. of ICASSP}.

\bibitem[\protect\citename{Yarowsky and Wicentowski}2000]{yarowsky:00}
David Yarowsky and Richard Wicentowski.
\newblock 2000.
\newblock Minimally supervised morphological analysis by multimodal alignment.
\newblock In {\em Proc. of ACL}.

\bibitem[\protect\citename{Zeiler}2012]{zeiler2012adadelta}
Matthew~D Zeiler.
\newblock 2012.
\newblock Adadelta: An adaptive learning rate method.
\newblock {\em arXiv preprint arXiv:1212.5701}.

\end{thebibliography}
\bibliographystyle{naaclhlt2016}

\end{document}